\documentclass[letterpaper]{article} 
\usepackage{aaai24}  
\usepackage{times}  
\usepackage{helvet}  
\usepackage{courier}  
\usepackage[hyphens]{url}  
\usepackage{graphicx} 
\usepackage{natbib}  
\usepackage{caption} 
\usepackage{algorithm}
\usepackage{algorithmic}

\usepackage{newfloat}
\usepackage{listings}
\DeclareCaptionStyle{ruled}{labelfont=normalfont,labelsep=colon,strut=off} 
\floatstyle{ruled}
\newfloat{listing}{tb}{lst}{}
\floatname{listing}{Listing}
\usepackage[utf8]{inputenc} 
\usepackage[T1]{fontenc}    
\usepackage{url}            
\usepackage{nicefrac}       
\usepackage{microtype}      

\usepackage{amsmath,amssymb,amsfonts}
\usepackage{algorithmic}
\usepackage{graphicx}
\usepackage{textcomp}
\usepackage{xcolor}
\usepackage{color}
\usepackage{float}
\usepackage{booktabs}
\usepackage{algorithmic}
\usepackage{subcaption}
\usepackage{multirow}
\usepackage{threeparttable}
\usepackage{xspace}
\usepackage[capitalise]{cleveref}

\usepackage{todonotes}
\usepackage{tablefootnote}
\usepackage{makecell}
\usepackage{picinpar}

\usepackage{color}
\usepackage{cleveref}

\newcommand{\ie}{{\textit{i.e.}},\xspace}
\newcommand{\eg}{{\textit{e.g.}},\xspace}
\newcommand{\btz}{\textit{resp.},\xspace}

\newcommand* \ourmethodname{\textit{DeepPointMap}\xspace}
\newcommand* \ExtractionNet{DPM Encoder\xspace}
\newcommand* \RegistrationNet{DPM Decoder\xspace}
\newcommand* \TheFirstHead{Similarity Head\xspace}
\newcommand* \TheSecondHead{Offset Head\xspace}
\newcommand* \TheThirdHead{Overlap Head\xspace}

\nocopyright
\title{DeepPointMap: Advancing LiDAR SLAM with Unified Neural Descriptors}
\author {
    Xiaze Zhang\textsuperscript{\rm 1},
    Ziheng Ding\textsuperscript{\rm 1},
    Qi Jing\textsuperscript{\rm 1}
    Yuejie Zhang\textsuperscript{\rm 1}
    Wenchao Ding*\textsuperscript{\rm 1}
    Rui Feng*\textsuperscript{\rm 1}
}
\affiliations {
    \textsuperscript{\rm 1}Fudan University\\
}

\usepackage{bibentry}

\begin{document}

\maketitle

\begin{abstract}
    Point clouds have shown significant potential in various domains, including Simultaneous Localization and Mapping (SLAM). However, existing approaches either rely on dense point clouds to achieve high localization accuracy or use generalized descriptors to reduce map size. Unfortunately, these two aspects seem to conflict with each other. To address this limitation, we propose a unified architecture, \ourmethodname, achieving excellent preference on both aspects.
    We utilize neural network to extract highly representative and sparse neural descriptors from point clouds, enabling memory-efficient map representation and accurate multi-scale localization tasks (\eg odometry and loop-closure).
    Moreover, we showcase the versatility of our framework by extending it to more challenging multi-agent collaborative SLAM. The promising results obtained in these scenarios further emphasize the effectiveness and potential of our approach.
\end{abstract}

\section{Introduction}
\label{sec:Introduction}

Simultaneous Localization and Mapping (SLAM) is a fundamental problem in robotics and autonomous driving, which aims to reconstruct the map of the explored environment while simultaneously estimating the location of the agent within it. It plays a vital role in enabling autonomous agents to navigate and understand their surroundings. Point clouds have gained prominence as a powerful representation for capturing intricate 3D structures of the environment. Taking advantage of this, LiDAR SLAM has become a prominent approach for achieving accurate localization and generating high-quality maps.

\begin{figure*}[t]
    \centering
    \includegraphics[width=.90\textwidth]{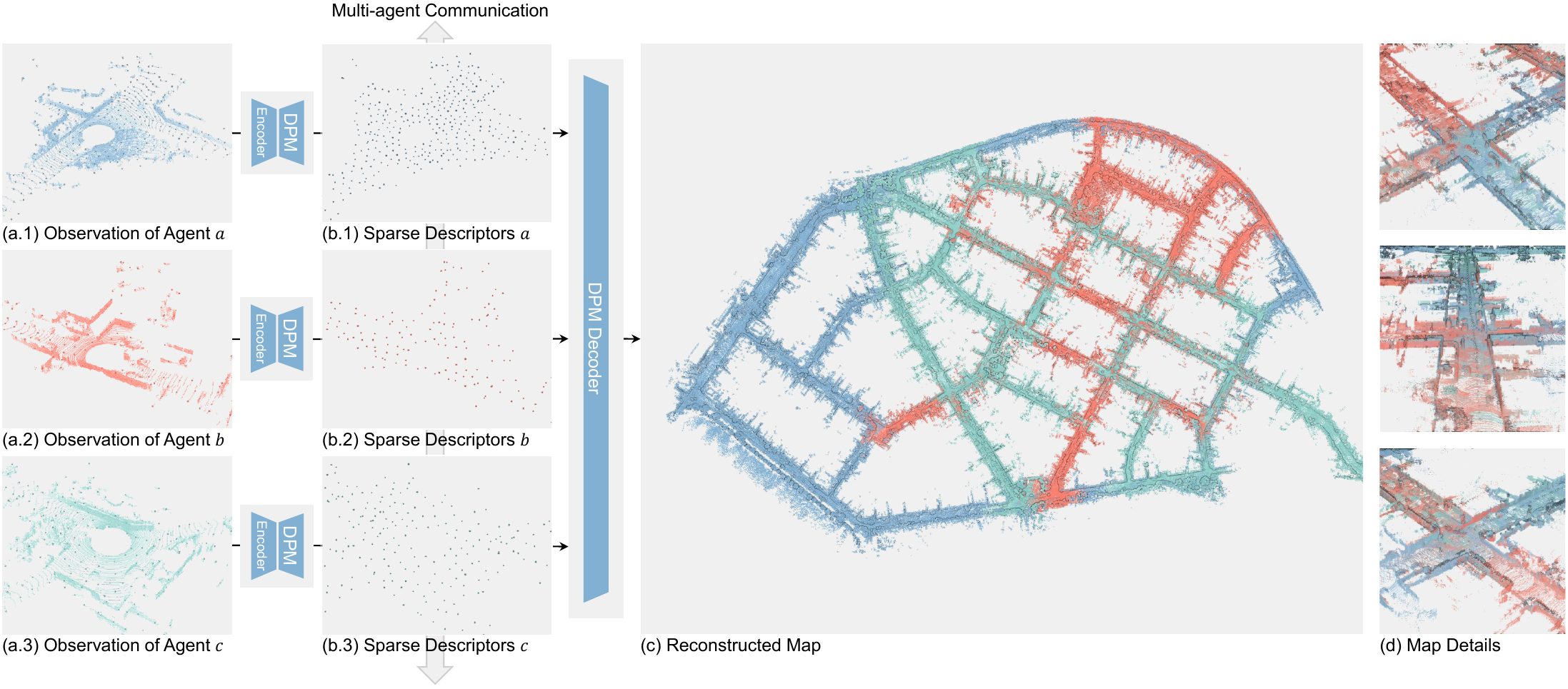}
    \caption{An illustration of the \ourmethodname. The agents collect point cloud (a) and extract it to sparse descriptors (b) locally. These descriptors are then gathered to reconstruct the complete map by multi-agent cooperative SLAM (c).} \label{fig:First Figute}
\end{figure*}

Currently, several SLAM methods~\cite{shan2018lego,chen2019suma++,pan2021mulls,vizzo2021poisson,dellenbach2022ct} based on hand-crafted features have demonstrated impressive performance on specific benchmarks. However, hand-crafted features are often either too sparse or over redundant, due to inefficacy in capturing semantic information. Sparse features may result in low-fidelity maps while dense features may be memory inefficient and problematic for large-scale reconstruction.
Some methods~\cite{qin2022geometric,yew2022regtr} use deep learning-based approaches to extract more compact features and achieve global registration. Although yielding promising results on point cloud registration benchmarks, they are hard to directly adapt to SLAM tasks since they can only deal with the registration task of single input size (\eg scan-level), and can be heavily impacted by cumulative errors.
Furthermore, most of the existing methods adopt pure geometric pipeline to compose the SLAM system, \eg involving hand-crafted feature extraction, iterative-based odometry and projection-based loop-closure. The geometric pipeline may lose considerable information from sensor inputs, which makes a neural information processing pipeline a more promising direction.

To tackle these challenges, we present \ourmethodname (DPM), a novel deep learning-based LiDAR SLAM framework including two neural networks: (1) the DPM Encoder which extracts unified neural descriptors to represent the environment efficiently, and (2) the \RegistrationNet which performs multi-scale matching and registration (\ie odometry and loop-closure) based on the aforementioned neural descriptors.
As illustrated in \cref{fig:First Figute}, the highly compact neural descriptors provide a novel solution for environment encoding and facilitate high-fidelity mapping and information sharing (\eg in multi-agent cooperative SLAM), while the \RegistrationNet renders a unified neural information processing pipeline for neural descriptors. 
Unlike other neural descriptor-based methods which can be only used in limited scenarios or specific SLAM components, our descriptor is used to accomplish several sub-tasks of SLAM task in a unified manner, yielding exceptional localization accuracy, memory efficiency, map fidelity, and real-time processing.

In summary, we introduce an efficient LiDAR SLAM framework predominantly based on neural networks, offering a promising solution for advancing the field of LiDAR SLAM. It achieves a new \textit{state-of-the-art} (SOTA) in localization accuracy, preserving a high-fidelity map reconstruction, with smaller memory consumption. These advantages can be further exploited in more challenging applications such as multi-agent cooperative SLAM which has limited communication bandwidth for feature sharing among agents.
Our contributions can be summarized as:
\begin{itemize}
    \item We propose using neural descriptors for online LiDAR SLAM. Compared to the traditional geometric descriptors, our neural descriptors serve as a compact map representation and facilitate a unified SLAM architecture, achieving desirable localization accuracy, memory efficiency, and map fidelity.
    \item We propose \ourmethodname\footnote{The source code of our approach will be made available in GitHub.}, a novel unified learning-based framework for LiDAR SLAM by conducting multi-scale matching and registration based on the aforementioned neural descriptors. The proposed framework achieves SOTA performance in localization accuracy and memory consumption, with real-time processing speed.
    \item To highlight the advantages of our framework, we further extend \ourmethodname to multi-agent cooperative SLAM task. Experimental results show that our approach can obtain better localization accuracy and mapping fidelity with constrained communication overhead.
\end{itemize}

\section{Related Work}\label{sec:Related Work}

\textbf{Map Reconstruction and Representation.}
Accurately and efficiently reconstructing the map of the environment is a crucial objective of LiDAR SLAM. Two of the most natural ways to represent the map are: (1) simply aligning the point clouds in global coordinates to construct a huge map-scale point cloud, or (2) storing the points explicitly into voxel-grid~\cite{dellenbach2022ct,vizzo2023kiss}.
However, these methods can only adequately describe the map through a substantial number of points, resulting in significant memory overhead. Meanwhile, they can not be easily modified once the mapping is completed.
Some methods use hand-crafted descriptors such as curvature, density and normal to represent the environment.  IMLS~\cite{deschaud2018imls} used \textit{implicit moving least squares surfaces}, while SuMa~\cite{behley2018efficient} and SuMa++~\cite{chen2019suma++} represented the point cloud into surfels. PUMA~\cite{vizzo2021poisson} introduced a surface mesh representation that better captured the geometric appearance of objects in the scene. However, these hand-crafted features are weakly representable, thus require complex registration algorithms to achieve accurate SLAM tasks.
Recently, some methods use neural networks to extract implicit features from point clouds. For instance, SegMatch~\cite{dube2017segmatch} and its subsequent work SegMap~\cite{dube2020segmap} extracted condensed segment-level features as descriptors for localization and map representation. Other approaches, such as~\cite{liu2019seqlpd, cattaneo2022lcdnet, xiang2022delightlcd, arce2023padloc}, extract global features of point clouds and predict the transformation based on these global features. However, these features tend to be either overly coarse or only applicable to specific pairs, limiting their universality for high-precision odometry.
Moreover, NeRF~\cite{xu2022pointnerf} also shows credible ability of representing point cloud. However, it usually has difficulties in online updating and accuracy localization, which is essential to autonomous driving-related tasks.
In contrast, our \ourmethodname utilizes sparse efficient neural descriptors to uniformly represent all point cloud scenarios with low memory overhead. These descriptors can be employed for various tasks in SLAM, including odometry and loop-closure, without any re-extraction.

\textbf{Point Cloud Localization.}
Localization is a fundamental task in LiDAR SLAM, encompassing scan-level localization (\ie odometry) and map-level localization (\ie loop-closure). Achieving accurate localization heavily relies on estimating the transformation between two point clouds.
Most traditional methods are based on the Iterative Closest Point (ICP) method~\cite{besl1992method} and its variants~\cite{low2004linear,censi2008icp,ramalingam2013theory}, which is a classic local registration algorithm with high computational complexity. To improve accuracy and efficiency, LOAM~\cite{zhang2014loam} and its derived algorithms~\cite{shan2018lego, wang2021f} detected edge and planar points as key points, which were used for more efficient numerical optimization in alignment. Additionally, MULLS~\cite{pan2021mulls} classified key points into more specific classes to establish more accurate correspondences.
However, these local registration methods are primarily limited to initial transformation guess (\eg from kinematic priors), which constrains their application to loop-closure and multi-agent SLAM without prior information.
With the introduction of PointNet~\cite{qi2017pointnet}, numerous deep learning methods for point cloud registration have emerged. Most of these methods~\cite{yuan2020deepgmr,huang2021predator,cao2021pcam,yew2022regtr,qin2022geometric,qin2023geotransformer} estimated correspondence in a global scope by training models to minimize feature distances between corresponding points. They then solved the transformation matrix using SVD~\cite{arun1987least}. DCP~\cite{wang2019deep} applied this strategy to loop-closure, introducing learning-based methods to the SLAM field.
However, these approaches are impractical for tackling both localization tasks in a unified manner as they can only handle single-scale inputs.
In contrast, our approach, \ourmethodname, based on unified descriptors, excels at global registration tasks involving multi-scale inputs, such as maps composed of multiple independent scans. This enables direct handling of various localization tasks, including scan-to-map odometry and loop-closure without any priors or iterations.

\textbf{Multi-agent Collaborative SLAM.}
Multi-agent SLAM requires collaboration among agents to improve mapping completeness and accuracy. Many approaches aim to reduce transmitted information to achieve efficient and reliable communication among multiple agents.
DDF-SAM~\cite{cunningham2010ddf} is an extended smoothing and mapping method for decentralized data fusion in multi-agent SLAM. \citet{lazaro2013multi} proposed a method based on condensed measurements~\cite{grisetti2012robust}, where agents only exchanged condensed measurements to enhance their trajectory estimation. These approaches primarily focus on reducing overhead or increasing robustness in communication, rather than reducing the size of descriptors.
In contrast, our approach focuses on fundamentally reducing communication overhead through lightweight and precise descriptors, making it widely applicable to both single- and multi-agent SLAM scenarios.

\section{Network Architecture}\label{sec:Network Architecture}

\begin{figure*}[htb]
    \centering
    \includegraphics[width=.90\textwidth]{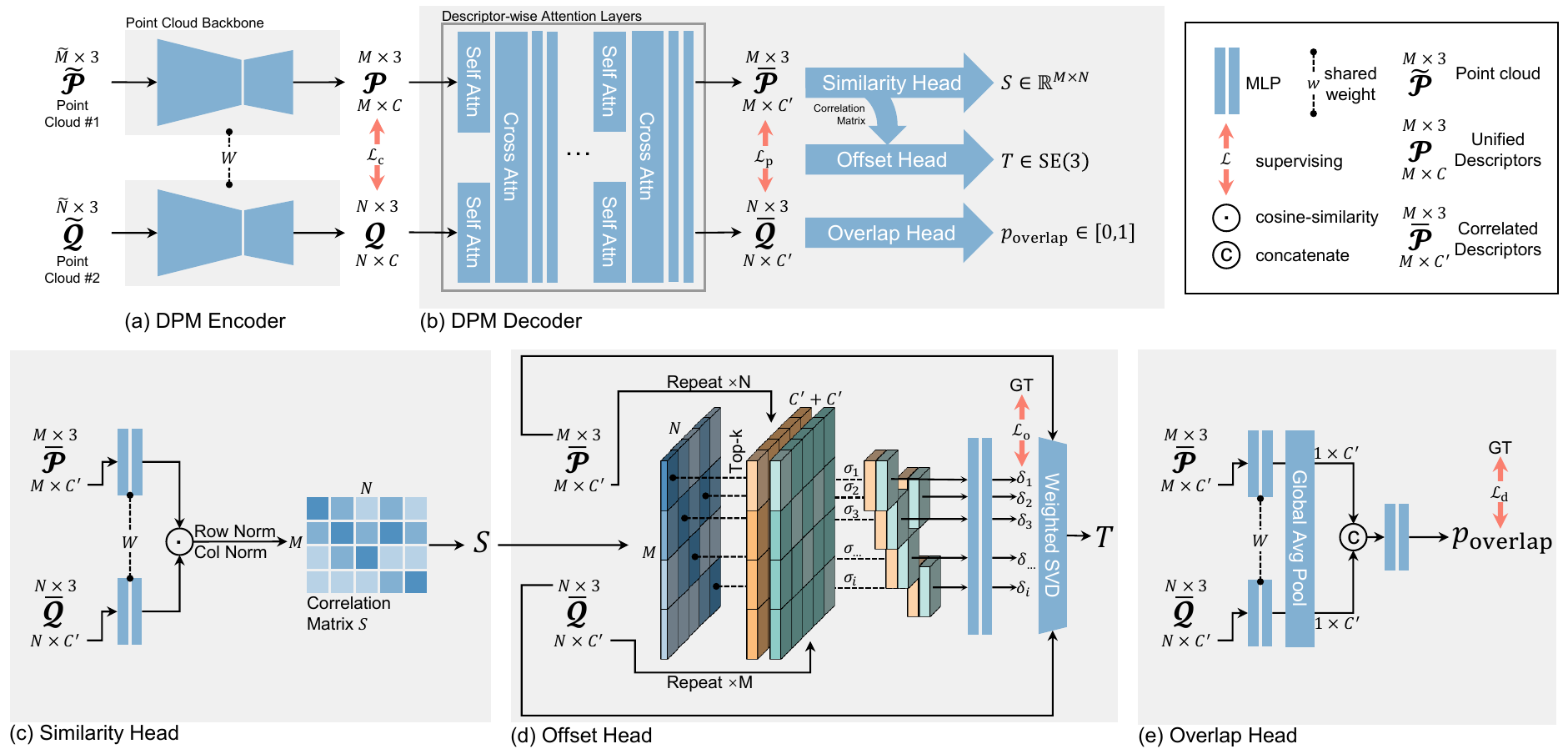}
    \caption{Overview of network architecture.} \label{fig:Overview of network architecture.}
\end{figure*}

The key step of \ourmethodname is to extract the unified descriptors and utilize them in localization tasks. As described in \cref{fig:Overview of network architecture.}, \ourmethodname consists of two parts: (a) \textbf{\ExtractionNet}: a point cloud backbone such as PointNeXt~\cite{qian2022pointnext}, and (b) \textbf{\RegistrationNet}: contains a Point-wise Attention Block and three heads as shown in \cref{fig:Overview of network architecture.} (c), (d) and (e).

\subsection{\ExtractionNet}
\label{subsec:Extraction Network}

\ourmethodname utilizes sparse descriptors that contain compressed semantic information to represent the complex point cloud for LiDAR SLAM.
A point cloud $\widetilde{\mathcal{P}}$, which can be represented as a set of 3D points $\widetilde{\mathcal{P}} = \left\{ \widetilde{\mathbf{p}}_1, \cdots, \widetilde{\mathbf{p}}_{\widetilde{N}} \right\} \in \mathbb{R}^{\widetilde{N}\times 3}$, is feeded into \ExtractionNet to extract sparse descriptors. The \ExtractionNet is essentially a PointNeXt~\cite{qian2022pointnext} backbone, which is one of the most famous neural architectures for point cloud understanding improved from PointNet~\cite{qi2017pointnet} and PointNet++~\cite{qi2017pointnet++}. The backbone samples sparse keypoints $\mathbf{p}_i^{\mathrm{xyz}} = \{x_i, y_i,z_i\} \in \mathbb{R}^3$ from dense point clouds $\widetilde{\mathcal{P}}$ and extract their neighboring geometric features $\mathbf{p}_i^{\mathrm{feat}} = \{f_i^1, \cdots, f_i^C\} \in \mathbb{R}^C$, where $C$ is the dimension of feature vector. The final output of the backbone, denoted by $\mathcal{P} = \left\{ \mathbf{p}_1,\mathbf{p}_2, \cdots,\mathbf{p}_M \right\} \in \mathbb{R}^{M\times (3+C)}$, is named as \textit{unified descriptor cloud} to represent the dense point cloud.

It is easy to merge multiple descriptor clouds by an union operation $\bigcup_i{\left\{ \mathcal{P}_i \right\}}$. In localization, we utilize this method to merge multiple scan-level descriptor clouds, thus forming the map-level descriptor clouds without any re-extraction.

\subsection{\RegistrationNet}
\label{subsec:Registration Network}

\RegistrationNet predicts the transformation between two aforementioned descriptor clouds, which consists of four parts: (1) Descriptor-wise Transformer Block, which is used for fusing deep correlation features between two input descriptor clouds, (2) Similarity Head, which aims to calculate the correspondence of descriptors between two clouds, (3) Offset Head, which predicts the relative positional offset between corresponding descriptors, enabling a more precise transformation estimation, and (4) Overlap Head, which predicts the probability of loop-closure occurred between two descriptor clouds, identifying potential loop closures in SLAM system.

\textbf{Descriptor-wise Transformer Block.}
As shown in \cref{fig:Overview of network architecture.} (b), \RegistrationNet takes a descriptor cloud pair $(\mathcal{P}, \mathcal{Q})$ as input and feeds them into the Descriptor-wise Transformer Block.
Inspired by \cite{yew2022regtr}, this block consists of three Descriptor-wise Transformer layers, where each layer comprises three sub-layers in sequence: (1) Multi-Head Self-Attention layer that operates independently on $\mathcal{P}$ and $\mathcal{Q}$, (2) Multi-Head Cross-Attention layer that fuses and exchanges contextual information between $\mathcal{P}$ and $\mathcal{Q}$, and (3) MLP that enhances the fitting capability.
All layers are weight-shared to reduce the parameter number. Residual connections and normalization are applied to expedite convergence and stabilize the gradient.
As the output of Descriptor-wise Transformer Block, the descriptor cloud pair is transformed into \textit{correlated descriptors}, denoted as $\bar{\mathcal{P}} = \left\{\bar{\mathbf{p}}_1, \cdots, \bar{\mathbf{p}}_M \right\} \in \mathbb{R}^{M\times (3+C^\prime)}$ (\btz $\bar{\mathcal{Q}}$). The position of correlated descriptor is unchanged in this process, \ie $\bar{\mathbf{p}}_i^\mathrm{xyz} \equiv \mathbf{p}_i^\mathrm{xyz}$.

\textbf{\TheFirstHead.}
The correlated descriptors are further projected into a new feature space using an MLP (denoted as $\mathrm{Head_{sim}}(\cdot)$). The correspondence matrix $S$ is then calculated by descriptor-wise cosine-similarity (denoted as $\odot$) $S_{i,j} = \mathrm{Head_{sim}}(\mathbf{p}_i^\mathrm{feat}) \odot \mathrm{Head_{sim}}(\mathbf{q}_j^\mathrm{feat})$. The higher similarity indicates that descriptor pair $\mathbf{p}_i^\mathrm{xyz}$ and $\mathbf{q}_j^\mathrm{xyz}$ are closer in global coordinate. When inferring, we select the top-$k$ descriptor pairs with the highest value $S_{i,j}$ to form the correspondence $\sigma = \arg\mathrm{top}_k{S}$.

\textbf{\TheSecondHead.}
Due to the sparsity of descriptors, the positions of corresponding descriptors are not likely to be perfectly coincide in global coordinate, which will introducing error to the transformation estimated by SVD even if the ground truth correspondences were given. To address this issue, we use \TheSecondHead to predict the relative positional offset $\delta_i \in \mathbb{R}^3$ between descriptor pairs $( \mathbf{\bar{p}}_i,\mathbf{\bar{q}}_i ) \in \sigma$, enabling us to achieve more accurate alignment.
It is necessary to mention that this prediction is directional: for the direction of $\bar{\mathcal{P}} \to \bar{\mathcal{Q}}$, we aim to find the offset ${\delta}_i^{\rightarrow}$ (\btz $\delta_i^{\leftarrow}$) that aligns $\mathbf{\bar{q}}_i$ to $\mathbf{\bar{p}}_i$ in $\bar{\mathcal{P}}$'s coordinate and vice versa.
We concatenate (denoted as $\oplus$) the features of the paired descriptors and use an MLP to predict the precise offset $\delta_i = \mathrm{Head_{offset}}( \mathbf{\bar{p}}^{\mathrm{feat}}_i \oplus \mathbf{\bar{q}}^{\mathrm{feat}}_i )$. After that, a weighted-SVD~\cite{arun1987least} is applied to solve the transformation $T$:
\begin{footnotesize}
    \begin{align}\label{eq:Solve Transformation with delta}
        \overset{\rightarrow}{\epsilon} & = \sum_i{\left\Vert  T \left(\mathbf{\bar{p}}_i^{\mathrm{xyz}} + \overset{\rightarrow}{\delta}_i\right) - \mathbf{\bar{q}}_i^{\mathrm{xyz}} \right\Vert_2^2} \\
        \overset{\leftarrow}{\epsilon}  & = \sum_j{\left\Vert  T \mathbf{\bar{p}}_j^{\mathrm{xyz}}  - \left( \mathbf{\bar{q}}_j^{\mathrm{xyz}} + \overset{\leftarrow}{\delta}_j\right)  \right\Vert_2^2} \\
        T                               & = \arg\min_{T}{\left( \overset{\rightarrow}{\epsilon} + \overset{\leftarrow}{\epsilon} \right)}
    \end{align}
\end{footnotesize}

\textbf{\TheThirdHead.}
The \TheThirdHead aims to predict whether the input descriptor cloud pair are overlapped, \ie their distance is shorter than a threshold $\varepsilon_{\mathrm{overlap}}=20\mathrm{m}$. This head has a concise structure as shown in \cref{fig:Overview of network architecture.} (e). The first share-weighted MLP applies nonlinear transformations to correlated descriptors, followed by an average pooling to gather global features of each cloud. The other MLP predicts the probability $p_\mathrm{overlap}$ of overlap based on the concatenated global features.


\subsection{Training}\label{subsec:Joint Training}

We jointly train the DPM Encoder and Decoder end-to-end with the following losses and strategies.

\textbf{Pairing Loss.}
We adopt InfoNCE loss~\cite{oord2018representation} as Pairing Loss $\mathcal{L}_{\mathrm{p}}$ on the correlated descriptors. For each $\bar{\mathbf{p}}_i \in \bar{\mathcal{P}}$, the descriptor $\mathbf{q}_j$ is assigned as either (1) \textit{positive} pair iff $j = \arg\min_{j} \left\Vert \mathbf{p}_i^\mathrm{xyz} - \mathbf{q}_j^\mathrm{xyz} \right\Vert_2^2 $ and $\left\Vert \mathbf{p}_i^\mathrm{xyz} - \mathbf{q}_j^\mathrm{xyz} \right\Vert_2 \le \varepsilon_{\mathrm{positive}}$, or (2) \textit{negative} pair. We denote these two pair sets as $\bar{\mathcal{Q}}_+$ and $\bar{\mathcal{Q}}_-$, respectively. The pairing loss for the correlated descriptors is:
\begin{footnotesize}
    \begin{equation}\label{eq:pairing loss}
        \mathcal{L}_{\mathrm{p}} =
        -\mathbb{E}_{\bar{\mathbf{p}}_i}
        \left[
            \log \left(
            \frac{
                \sum_{ \mathbf{\bar{q}}_j^{\mathrm{feat}} \in \bar{\mathcal{Q}}_+}\exp \left( \mathbf{\bar{p}}_i^{\mathrm{feat}} \odot \mathbf{\bar{q}}_j^{\mathrm{feat}} \right) / \tau
            }{
                \sum_{ \mathbf{\bar{q}}_j^{\mathrm{feat}} \in \bar{\mathcal{Q}}}\exp \left( \mathbf{\bar{p}}_i^{\mathrm{feat}} \odot \mathbf{\bar{q}}_j^{\mathrm{feat}}  \right) / \tau
            }
            \right)
            \right]
    \end{equation}
\end{footnotesize}

\textbf{Coarse Pairing Loss.}
In order to ensure the unified descriptor clouds extracted by \ExtractionNet can preliminary distinguish the correspondence of descriptors, we adopt the Coarse Pairing Loss $\mathcal{L}_{\mathrm{c}}$ similar to $\mathcal{L}_{\mathrm{p}}$ on unified descriptors $\mathcal{P}$ and $\mathcal{Q}$. We keep the definition of \textit{positive} pair as above, but split the \textit{negative} pair $\mathbf{q}_j$ as: (1) \textit{negative} pair iff $\left\Vert \mathbf{p}_i - \mathbf{q}_j \right\Vert_2 > \varepsilon_{\mathrm{positive}}$, otherwise (2) \textit{neutral} pair, which is not used to contribute the loss.
This strategy provides more inclusiveness for those pairs of descriptors that are close but not closest, because of the probability that these pairs will act as positive and negative pairs respectively at different times, thus causing ambiguity. We denote \textit{positive}, \textit{negative} and \textit{neutral} pairs as $\mathcal{Q}_+$, $\mathcal{Q}_-$ and $\mathcal{Q}_\circ$. The Coarse Pairing Loss is defined as:
\begin{footnotesize}
    \begin{equation}\label{eq:contrastive loss}
        \mathcal{L}_{\mathrm{c}} =
        -\mathbb{E}_{\mathbf{p}_i}
        \left[
            \log \left(
            \frac{
                \sum_{\mathbf{q}_j \in \mathcal{Q}_+}\exp \left( \mathbf{p}_i^{\mathrm{feat}} \odot \mathbf{q}_j^{\mathrm{feat}} \right)/ \tau
            }{
                \sum_{\mathbf{q}_j \in \mathcal{Q}_+ \cup \mathcal{Q}_-}\exp \left( \mathbf{p}_i^{\mathrm{feat}} \odot \mathbf{q}_j^{\mathrm{feat}} \right)/ \tau
            }
            \right)
            \right]
    \end{equation}
\end{footnotesize}

\textbf{Offset Loss.} We adapt the Offset Loss to train \TheSecondHead. Following the definition of three pair types above, we use both \textit{positive} and \textit{nurture} pairs to train the \TheSecondHead to predict the offsets.
\begin{footnotesize}
    \begin{equation}
        \mathcal{L}_{\mathrm{o}} =
        \mathbb{E}_{\mathbf{p}_i}
        \left[
            \frac{1}{\left\vert \bar{\mathcal{Q}}_+ \cup \bar{\mathcal{Q}}_\circ \right\vert}
            \sum _{\bar{\mathbf{q}}_j \in \bar{\mathcal{Q}}_+ \cup \bar{\mathcal{Q}}_\circ}
            \left\Vert
            \delta_{i,j} - \delta_{i,j}^\ast
            \right\Vert_\Sigma
            \right]
    \end{equation}
\end{footnotesize}
where $\delta_{i,j} = \mathrm{Head_{offset}}(\bar{\mathbf{p}}_i^\mathrm{feat}, \bar{\mathbf{q}}_j^\mathrm{feat})$ represents the predicted offset between $(\bar{\mathbf{p}}_i^\mathrm{xyz}, \bar{\mathbf{q}}_j^\mathrm{xyz})$, $\delta_i^\ast$ is the ground-truth and $\left\Vert \cdot \right\Vert_\Sigma$ is the Mahalanobis distance between them.
We utilize both \textit{positive} and \textit{neutral} pairs to stabilize the gradient during training thus accelerating the convergence and improving the robustness.
We use a different distance threshold $\varepsilon_{\mathrm{offset}}$ here to define \textit{positive} and \textit{negative} pairs.

\textbf{Overlap Loss.} We use Binary Cross Entropy (BCE) loss $\mathcal{L}_\mathrm{d}$ to train the \TheThirdHead. However, it will not be applied together with other losses.

\textbf{Overall Loss.}
The directional (\ie $\mathcal{P} \to \mathcal{Q}$) loss of \ourmethodname is defined in \cref{eq:overall loss}. We use the average of loss in both directions, with the hyper-parameters of $\varepsilon_{\mathrm{positive}}=1\mathrm{m}$, $\varepsilon_{\mathrm{offset}}=2\mathrm{m}$, $\tau=0.1$, $\lambda_1, \lambda_2, \lambda_3 = \left(1.0,0.1,1.0\right)$.
\begin{small}
    \begin{equation}
        \label{eq:overall loss}
        \begin{aligned}
            \mathcal{L}_{\mathrm{r}}=
            \lambda_1 \mathcal{L}_{\mathrm{p}} +
            \lambda_2 \mathcal{L}_{\mathrm{c}} +
            \lambda_3 \mathcal{L}_{\mathrm{o}}
        \end{aligned}
    \end{equation}
\end{small}

\textbf{Data Augmentation and Curriculum Learning.}
In some cases, vehicles nearby may obstruct the LiDAR's scan, which results in large point vacuum areas and may reduce localization accuracy.
Therefore, we propose a novel data augmentation named \textit{Random Occlusion} to simulate this situation. We first randomly generate some virtual boxes with random size and position. Each box introduces an occlusion effect by removing all points that pass through it. This augmentation can significantly improve the performance on occlusion cases.
In our SLAM task, both scan- and map-level registrations are involved. However, regular training strategies~\cite{yew2022regtr, qin2023geotransformer} only focus on scan-to-scan, \ie fixed-scale samples, which neglect multi-scale registration. To address this, we adapt the curriculum learning strategy to progressively train \ourmethodname from simple to complex scenarios.
In our training procedure, we gradually increase the scale of the descriptor clouds, which leads the model to gradually learn the capability of large-scale registration tasks. After that, we freeze all modules except \TheThirdHead and perform an individual training procedure to train \TheThirdHead. The scan pair is sampled with an equal probability of positive and negative samples. More describes can be found in \textit{supplementary material}.

\section{ \ourmethodname Framework}\label{sec:The DeepPointMap Framework}

During inference, we compose the network into a full SLAM system, named \ourmethodname. We introduce it in two aspects: (1) Mapping, including map reconstruction and optimization, and (2) Localization, containing odometer and loop-closure. In addition, we further briefly introduce a multi-agent expansion of \ourmethodname.

\textbf{Pose-Graph based Map.} We adopt \textit{Pose-Graph} to represent our reconstructed map, denoted as $\mathcal{M} = \left(V, E\right)$. Each observation $v_t \in V$ at time-step $t$ contains the estimated pose $T_t$ of the agent and the corresponding descriptor cloud $\mathcal{P}_t$. Meanwhile, the edge $e(v_i, v_j) \in E$ represents the positional relation between observations $v_i, v_j$.
To ensure global consistency in the reconstructed map, we utilize standard pose graph optimization to globally optimize the pose estimates $T_t$ of all observations $v_t$ once the loop-closure (see description below) edge is inserted.

\textbf{Keyframe Selection.}
The main limitation of pose-graph based map representation is that agent seems to continuously insert new observations into the map, regardless of whether the location has been visited and already well-represented.
To address this issue, we adopt a simple keyframe selection algorithm. For a new observation $v_x$, after estimating its pose $T_x$ (described below), and accept $v_x$ and insert it into $\mathcal{M}$ iff the distance of the nearest keyframe is greater than a dynamic threshold $\varepsilon_\mathrm{keyframe}$, otherwise $v_x$ is discarded. Threshold $\varepsilon_\mathrm{keyframe}$ will increase if the current registration confidence is low and vise versa.

\textbf{Odometer.} The odometer estimates the current pose of the agent by continuously predicting the transformation $T_{x,t}$ between the current observation $v_t$ and a previous one $v_i$. Unlike temporal-based method, we first search the \textit{K}-nearest-neighbors $N(v_{t-1})$ of $v_{t-1}$ and find the closest keyframe $v_i$ respect to the last known position. Then, $T_t$ is estimated using \RegistrationNet by registering the descriptor clouds $\mathcal{P}_t$ and $\mathcal{P}_i$. To obtain a more accurate pose estimation $T_{i,t}$, we apply a scan-to-map registration between $\mathcal{P}_t$ and $\bigcup{N(v_{i})}$ as soon as the observation $v_t$ is accepted (as described above). Finally, we insert an \textit{odometer edge} $e(v_i,v_t)$ to the pose-graph.

\textbf{Loop-Closure.}
The loop-closure procedure is composed of three stages: detection, registration, and verification.
Given an observation $v_t$, we apply a spatial search to find the candidate observation list $V_c \subset V$ within $100\mathrm{m}$ with respect to $v_t$. For each candidate $v_i \in V_c$, \TheThirdHead predicts their loop probability $p_\mathrm{overlap}$. If the probability exceeds a threshold $\varepsilon_\mathrm{loop}$, the loop proposal is accepted.
Then, we adapt a map-to-map registration using \RegistrationNet on descriptor clouds $\bigcup{N(v_{t})}$ and $\bigcup{N(v_{i})}$, to estimate an accurate pose $T_{t, i}$.
Finally, a \textit{loop-edge} $e(v_i,v_t)$ is inserted into the pose-graph.

\textbf{Multi-Agent Cooperative SLAM.}
In our multi-agent \ourmethodname framework, each agent maintains its own SLAM system and performs odometry and loop-closure locally. To ensure global consistency of trajectories and to merge the observations from each agent, we extend the \ourmethodname framework as follows:
(1) Map Representation: Unless the agents' trajectories are spatially intersected, the map $\mathcal{M}$ is separated into multiple components $\mathcal{M}^1, \cdots \mathcal{M}^m$, where each component $\mathcal{M}^a=(V^a, E^a)$ represents a sub-map reconstructed by agent $a$. 
(2) Trajectory Merging:
To detect trajectory intersections, we use the same strategy as single-agent loop detection, except assigning all observations from other agents as the candidate list (if no additional global position \eg GNSS is provided). When a trajectory intersection between observations $v_i^a$ and $v_j^b$ is verified, a \textit{cross edge} $e(v_i^a, v_j^b)$ is inserted into the pose-graph that connects two components. Subsequently, a pose-graph optimization is applied to align trajectories.

\section{Experimental Analysis}
\label{sec:Experimental Result}

We conduct extensive experiments to evaluate our proposed \ourmethodname with respect to current \textit{state-of-the-art} algorithms on benchmark datasets.

\textbf{Experimental Settings.}
We train the network following the strategy described in \cref{subsec:Joint Training}, with AdamW optimizer~\cite{2019adamW}, initial learning rate $lr=1e-3$, weight decay $wd=1e-4$, cosine lr scheduler, on 6$\times$ RTX 3090 GPU. For all tasks, the network is trained for 12 epochs. When training \TheThirdHead, we decay $lr$ and $wd$ with a rate of 0.1. We use the model from the last epoch for evaluation. More training and inference details including data preprocessing can be found in \textit{supplementary material}.

\textbf{Benchmarks.} We conducted the experiments on the following four autonomous driving-oriented datasets: 
(1) SemanticKITTI~\cite{behley2019semantickitti,behley2021towards} is a widely used benchmark dataset based on KITTI Vision Benchmark~\cite{geiger2012we}, consisting of 11 sequences (\texttt{00}-\texttt{10}) of LiDAR scans collected in various driving scenarios. We use the first 6 sequences as training-set.
(2) KITTI-360~\cite{liao2022kitti} includes 11 large-scale sequences (\texttt{00},\texttt{01},\texttt{03}-\texttt{10} and \texttt{18}). We use the first 6 sequences as training-set.
(3) MulRan~\cite{gskim-2020-mulran} is a range dataset for urban place recognition and SLAM studies, which contains 12 sequences collected from 4 different environments. Following ~\citet{kim2021scan}, we use \texttt{KAIST03} and \texttt{Riverside02} sequences as testing-set, and use all 6 sequences collected in \texttt{Sejong} and \texttt{DCC} as training-set.
(4) KITTI-Carla~\cite{deschaud2021kitti} is a synthetic dataset with 7 sequences (\texttt{Town01}-\texttt{Town07}) generated on the CARLA simulation platform~\cite{Dosovitskiy17}, which provides noise-free data. We use all sequences in KITTI-Carla for training.
Unless specified, all other sequences are used as testing sets. The split of training and testing set is based on the frame amount of \textit{approx.} 6:4, without any manual picking.
For quantitative evaluations, we adapted the stranded Absolute Pose Error (APE) to evaluate the global accuracy of predicted trajectory, which is also used in KITTI-360 benchmark~\cite{liao2022kitti}.

\subsection{Localization Accuracy}
\label{subsec:Comparison between DeepPointMap and State-of-the-art Algorithms}

The experiment is designed to show the localization accuracy of our method. To this end, we compare our method with 6 recent \textit{state-of-the-art} Odometer and SLAM methods: KISS-ICP~\cite{vizzo2023kiss}, LeGO-LOAM~\cite{shan2018lego} and its following work SC-LeGO-LOAM~\cite{kim2021scan}, MULLS~\cite{pan2021mulls}, CT-ICP~\cite{dellenbach2022ct}, as well as a \textit{state-of-the-art} point cloud registration method GeoTransformer~\cite{qin2023geotransformer}.
To make a fair comparison, we execute their open-source code on our platform and evaluate them using the same settings.

\begin{figure*}[ht]
    \centering
    \includegraphics[width=.95\textwidth]{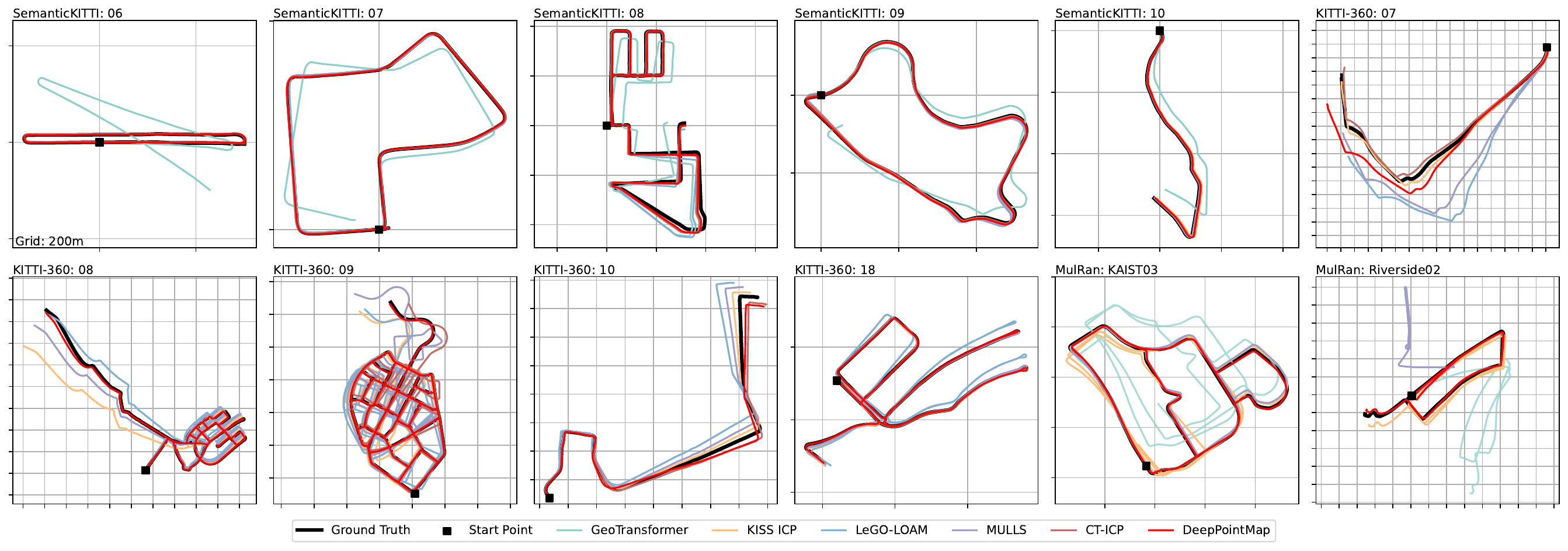}
    \caption{Trajectories Estimated with \ourmethodname and Comparison Methods on Testing Sequences.} \label{fig:Trajectories Estimated KITTI360}
\end{figure*}

\cref{fig:Trajectories Estimated KITTI360} presents the trajectories estimated by \ourmethodname vs. comparison methods on SemanticKITTI, KITTI-360 and MulRan. We observe that most methods achieve similar localization accuracy on SemanticKITTI, even for those approaches without global optimization. 
The performance difference between methods gradually emerges as the map scaled up. The KITTI-360 \texttt{09} is one of the longest sequences in our experiment, exceeding $10,000\mathrm{m}$ length and covering \textit{approx.} $900,000\mathrm{m}^2$ area. Most of the comparison methods fail to reconstruct a consistent global map.
However, \ourmethodname with high accuracy and robustness successfully manage to reconstruct it with very slight distortion.
For sequence \texttt{Riverside02}, some traditional methods struggled to produce accurate maps due to the scarcity of distinct reference objects in the scans. However, thanks to its superior feature extraction and representation capabilities, \ourmethodname successfully reconstructed these challenging scenes.

\begin{table*}[ht]
    \caption{Localization Accuracy of \ourmethodname and Other SOTA Methods (APE$\downarrow$).}
    \label{Table:Localization Accuracy of DeepPointMap and Other SOTA Methods}
    \begin{center}
        \resizebox{.95\textwidth}{!}{
            \begin{threeparttable}
                \begin{tabular}{l | rrrrr rrrrr rr}
                    \toprule[0.75pt]
                    \multicolumn{1}{c}{}                                 & \multicolumn{5}{c}{\textbf{SemanticKITTI}} & \multicolumn{5}{c}{\textbf{KITTI-360}} & \multicolumn{2}{c}{\textbf{MulRan}}                                                                                                                                                            \\
                    \cmidrule[0.5pt](lr){2-6} \cmidrule[0.5pt](lr){7-11} \cmidrule[0.5pt](lr){12-13}
                    \textbf{Method}                                      & \texttt{06}                                & \texttt{07}                            & \texttt{08}                         & \texttt{09}   & \texttt{10}   & \texttt{07}    & \texttt{08}   & \texttt{09}   & \texttt{10}   & \texttt{18}   & \texttt{KAIST03} & \texttt{Riverside02} \\
                    \midrule[0.5pt]
                    LeGO-LOAM~\cite{shan2018lego}                        & 0.88                                       & 0.67                                   & 8.84                                & 1.95          & 1.37          & 82.84          & 32.79         & 7.36          & 26.56         & 2.80          & failed            & failed                \\
                    SC-LeGO-LOAM~\cite{kim2021scan}                      & 1.02                                      & 1.46                                  & 6.23                               & 8.31         & 1.69         & 47.78          & 8.47         & 22.38         & 9.57         & 6.27         & 3.85            & 28.16                \\
                    MULLS~\cite{pan2021mulls}                            & \textbf{0.48}                              & 0.38                                   & 4.16                                & 1.99          & 0.97          & 47.25          & 8.24          & 93.88         & 11.99         & 1.52          & 6.63             & failed                \\
                    CT-ICP~\cite{dellenbach2022ct}                       & 0.56                                       & 0.43                                   & 4.07                                & 1.29          & 0.94 & \textbf{14.43} & -             & 11.41         & \textbf{7.11} & -             & -                & -                    \\
                    GeoTransformer~\cite{qin2023geotransformer}\tnote{1} & 24.38                                      & 8.71                                   & 22.68                               & 22.14         & 16.36         & 647.93         & 100.16        & 75.93         & 111.70        & 34.79         & 71.95            & 232.86               \\
                    KISS-ICP~\cite{vizzo2023kiss}                        & 0.61                                       & 0.35                                   & 3.58                                & 1.32          & \textbf{0.94}          & 20.51          & 26.24         & 24.00         & 8.75          & 2.22          & 12.85            & 27.16                \\
                    \midrule[0.5pt]
                    \ourmethodname (ours)                                & 0.92                                       & \textbf{0.27}                          & \textbf{3.44}                       & \textbf{1.27} & 1.28          & 77.82          & \textbf{5.26} & \textbf{1.02} & 10.47         & \textbf{1.04} & \textbf{1.68}             & \textbf{17.48}                \\
                    \ourmethodname (ours)\tnote{2}                       & 0.98                                       & 0.51                                   &   6.06                                & 1.38          & 1.73          & -              & -             & -             & -             & -             & -            & -                \\
                    \bottomrule[0.75pt]
                \end{tabular}
                \begin{tablenotes}
                    \footnotesize
                    \item[1] GeoTransformer is a point cloud registration method. We used to estimates trajectory by scan-to-scan registration.
                    \item[2] Cross-dataset transfer experiment of \ourmethodname, trained on KITTI-360 and KITTI-Carla, tested on unseen SemanticKITTI.
                \end{tablenotes}
            \end{threeparttable}
        }
    \end{center}
\end{table*}

\cref{Table:Localization Accuracy of DeepPointMap and Other SOTA Methods} reports the quantitative results, where \ourmethodname competes very favorably with its 6 peers on 12 sequences, by attaining 8 best results. Some results of CT-ICP are missing because CT-ICP is only able to run on its own provided data, and these missing sequences are not available to us. We also conduct a transfer experiment, as shown in the last line, to demonstrate the generalization ability on unseen data. We train \ourmethodname on KITTI-360 and KITTI-Carla and directly evaluate its performance on SemanticKITTI, which still shows favorable localization accuracy.

\subsection{Memory Efficiency}\label{subsec:Memory Efficiency}
\begin{figure}[H]
    \includegraphics[width=.95\columnwidth]{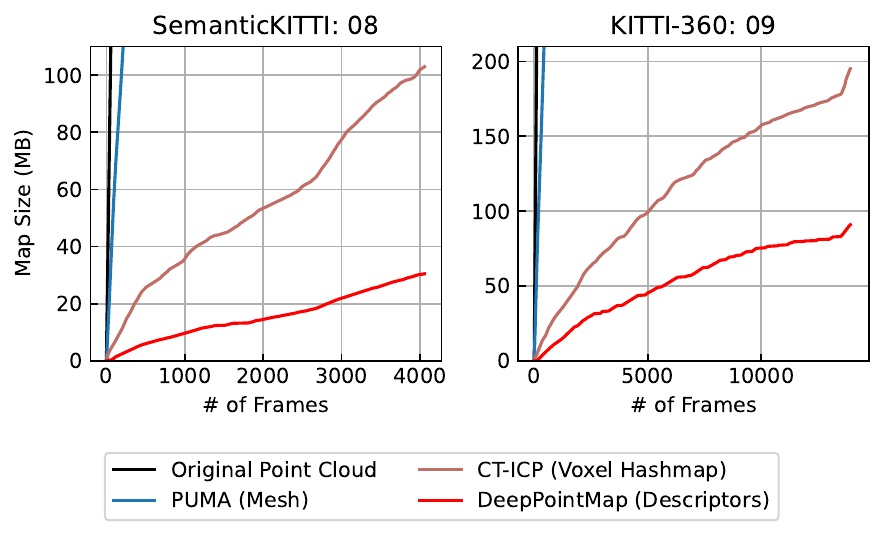}
    \caption{Map Size for Different Representations.}\label{fig:Map size compaired to other methods}
\end{figure}

Efficient map representation is crucial for reconstructing large-scale maps. To evaluate the memory efficiency of \ourmethodname, we compared the reconstructed map size with several widely used methods: Original Point Cloud, Voxel Hashmap~\cite{dellenbach2022ct} and Mesh~\cite{vizzo2021poisson}. To achieve a fair comparison, we only count the size of data that is necessary in further localization. All data is stored in \texttt{float32} dtype. \cref{fig:Map size compaired to other methods} shows the memory consumption for different map representations in two sequences.
As shown in \cref{fig:Map size compaired to other methods}, our \ourmethodname manages to save up to 70\% memory in SemanticKITTI \texttt{07} and about 50\% memory in KITTI-360 \texttt{08}, compared to other methods.

\subsection{Multi-agent Cooperative SLAM Experiment}
\label{subsec:Multi-agent Cooperative SLAM Experiment}

\begin{figure}[t]
    \centering
    \includegraphics[width=.95\columnwidth]{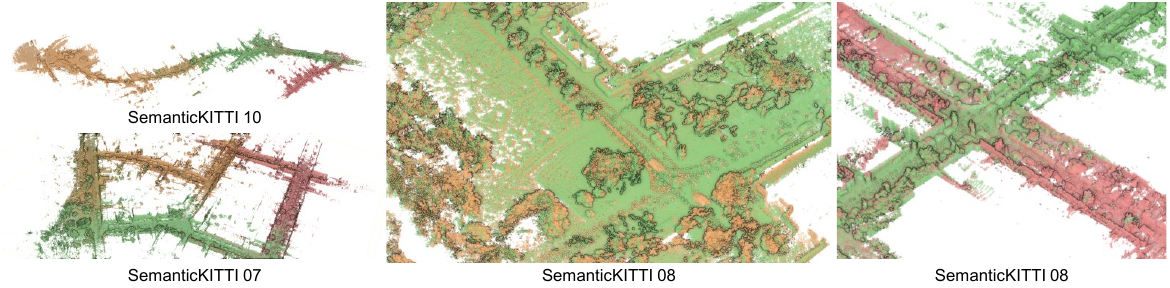}
    \caption{Multi-agent Cooperative Mapping and Localization.}\label{fig:Multi-agent Cooperative Mapping and Localization}
\end{figure}

To demonstrate the superiority of \ourmethodname, we extend our method to the multi-agent cooperative SLAM task described in \cref{sec:The DeepPointMap Framework}. We select 3 sequences from SemanticKITTI dataset and split them into 3 subsequences. The subsequences are then assigned to 3 individual agents to simulate the real-world multi-agent cooperative SLAM scenario. The resulting reconstructed point cloud is shown in \cref{fig:Multi-agent Cooperative Mapping and Localization}, where the observed point clouds of each agent are marked with different colors. Agents successfully recognize the trajectory-intersect and perform multi-agent loop-closure, thus successfully reconstructing the environment. More experiential results can be found in \textit{supplementary material}.

\subsection{Ablation Study}
\label{subsec:Ablation Study}

\begin{table}[!h]
    \caption{Ablation Study on SemanticKITTI.}\label{tab:Ablation Study Result}
    \begin{center}
        \resizebox{.80\columnwidth}{!}{
            \begin{threeparttable}
                \begin{tabular}{l|ccc}
                    \toprule[0.75pt]
                    \textbf{Method}                  & \texttt{08}    & \texttt{09} & \texttt{10} \\
                    \midrule[0.5pt]
                    \ourmethodname  (baseline)       & 4.7            & 2.53        & 1.35        \\
                    \textit{w/o} \TheSecondHead      & \makecell{3.69                             \\\scriptsize{(-21.5\%)}}        &    \makecell{2.84\\ \scriptsize{(+12.3\%)}}&   \makecell{63.26\\  \scriptsize{(failed)}}     \\
                    \textit{w/o} Coarse Pairing Loss & \makecell{3.92                             \\\scriptsize{(-16.6\%)} }    &      \makecell{2.58\\    \scriptsize{(+2.0\%)}}     &    \makecell{2.27\\ \scriptsize{(+68.1\%)}}     \\
                    \textit{w/o} \makecell{Curriculum Learning                                    \\ \& Random Occlusion} & \makecell{174.57                             \\\scriptsize{(failed)}  }      &  \makecell{92.61\\    \scriptsize{(failed)}}        &  \makecell{162.83\\ \scriptsize{(failed)} }      \\
                    \bottomrule[0.75pt]
                \end{tabular}
            \end{threeparttable}
        }
    \end{center}
\end{table}

To gain a deeper understanding of \ourmethodname, we conduct several ablation studies to investigate the role of various components. All variants, as well as the baseline, are trained on SemanticKITTI, using the training set of the first 8 sequences and the testing set of the last 3 sequences.

We remove \textit{\TheSecondHead}, \textit{Coarse Pairing Loss} and \textit{Curriculum Learning \& Random Occlusion} respectively to evaluate their importance.
Our training strategy plays a critical role as it enabled the model to learn the accurate and multi-scale representation of the environment.
\TheSecondHead was found to be necessary for achieving higher precision registration. It effectively mitigated the negative impact of descriptor sparsity and reduced the occurrence of trajectory prediction failures.
Coarse Pairing Loss helps the model extract features at an earlier stage. This facilitated the \RegistrationNet in capturing their fine-grained relationships and accelerated convergence, leading to improved performance of \ourmethodname.

\section{Conclusions}\label{sec:Conclusions}
We present \ourmethodname, a novel deep learning algorithm for LiDAR SLAM. This approach achieves accurate localization and reconstructs lightweight maps using uniform and efficient descriptors. Additionally, it demonstrates adaptability to multi-agent cooperative SLAM scenarios. Our method outperforms previous approaches on challenging benchmarks, particularly in large and complex urban scenes.

\textbf{Limitation.} Compared to traditional methods, neural network-based approaches require more precise data labels \ie the pose of each LiDAR scan. However, in the context of autonomous driving SLAM, label quality often falls short, potentially impacting experimental results. Notably, our method showed weaker performance in certain distant countryside sequences (like KITTI-360 \texttt{07}, \texttt{10}, etc.), due to sparse reference objects and limited geometric information in the point cloud. This could explain why our model lagged behind those utilizing kinematic estimation (\eg \citet{vizzo2023kiss}). To address this, we plan to incorporate visual modality in our future work.

\bibliography{aaai24}

\end{document}